\documentclass[10pt,twocolumn,letterpaper]{article}

\usepackage{cvpr}
\usepackage{times}
\usepackage{epsfig}
\usepackage{graphicx}
\usepackage{amsmath}
\usepackage{amssymb}
\usepackage{multirow}
\usepackage{authblk}

\usepackage[pagebackref=true,breaklinks=true,letterpaper=true,colorlinks,bookmarks=false]{hyperref}

\cvprfinalcopy 

\def\cvprPaperID{3980} 

\ifcvprfinal\fi
\begin{document}

\title{A Twofold Siamese Network for Real-Time Object Tracking}
\author[$\dagger$]{Anfeng He\thanks{This work is carried out while Anfeng He is an intern in MSRA.}}
\author[$\ddagger$]{Chong Luo}
\author[$\dagger$]{Xinmei Tian}
\author[$\ddagger$]{Wenjun Zeng}
\affil[$\dagger$]{CAS Key Laboratory of Technology in Geo-Spatial Information Processing and Application System, University of Science and Technology of China, Hefei, Anhui, China}
\affil[$\ddagger$]{Microsoft Research, Beijing, China}
\affil[ ]{\tt\small heanfeng@mail.ustc.edu.cn, \{cluo, wezeng\}@microsoft.com, xinmei@ustc.edu.cn}

\maketitle

\begin{abstract}
Observing that \underline{S}emantic features learned in an image classification task and \underline{A}ppearance features learned in a similarity matching task complement each other, we build a twofold \underline{Siam}ese network, named \underline{SA-Siam}, for real-time object tracking. 
SA-Siam is composed of a semantic branch and an appearance branch. Each branch is a similarity-learning Siamese network. 
An important design choice in SA-Siam is to separately train the two branches to keep the heterogeneity of the two types of features.
In addition, we propose a channel attention mechanism for the semantic branch. Channel-wise weights are computed according to the channel activations around the target position.
While the inherited architecture from SiamFC \cite{SiamFC} allows our tracker to operate beyond real-time, the twofold design and the attention mechanism significantly improve the tracking performance. The proposed SA-Siam outperforms all other real-time trackers by a large margin on OTB-2013/50/100 benchmarks. 

\end{abstract}

\section{Introduction}
\label{sec:introduction}

Visual object tracking is one of the most fundamental and challenging tasks in computer vision. Given the bounding box of an unknown target in the first frame, the objective is to localize the target in all the following frames in a video sequence. While visual object tracking finds numerous applications in surveillance, autonomous systems, and augmented reality, it is a very challenging task. For one reason, with only a bounding box in the first frame, it is difficult to differentiate the unknown target from the cluttered background when the target itself moves, deforms, or has an appearance change due to various reasons. For another, most applications demand real-time tracking. It is even harder to design a real-time high-performance tracker.

The key to design a high-performance tracker is to find expressive features and corresponding classifiers that are simultaneously \emph{discriminative} and \emph{generalized}. Being discriminative allows the tracker to differentiate the true target from the cluttered or even deceptive background. Being generalized means that a tracker would tolerate the appearance changes of the tracked object, even when the object is not known a priori. Conventionally, both the  discrimination and the generalization power need to be strengthened through online training process, which collects target information while tracking. However, online updating is time consuming, especially when a large number of parameters are involved. It is therefore very crucial to balance the tracking performance and the run-time speed. 


\begin{figure}[t!]
    \begin{center}
    \includegraphics[width=\columnwidth]{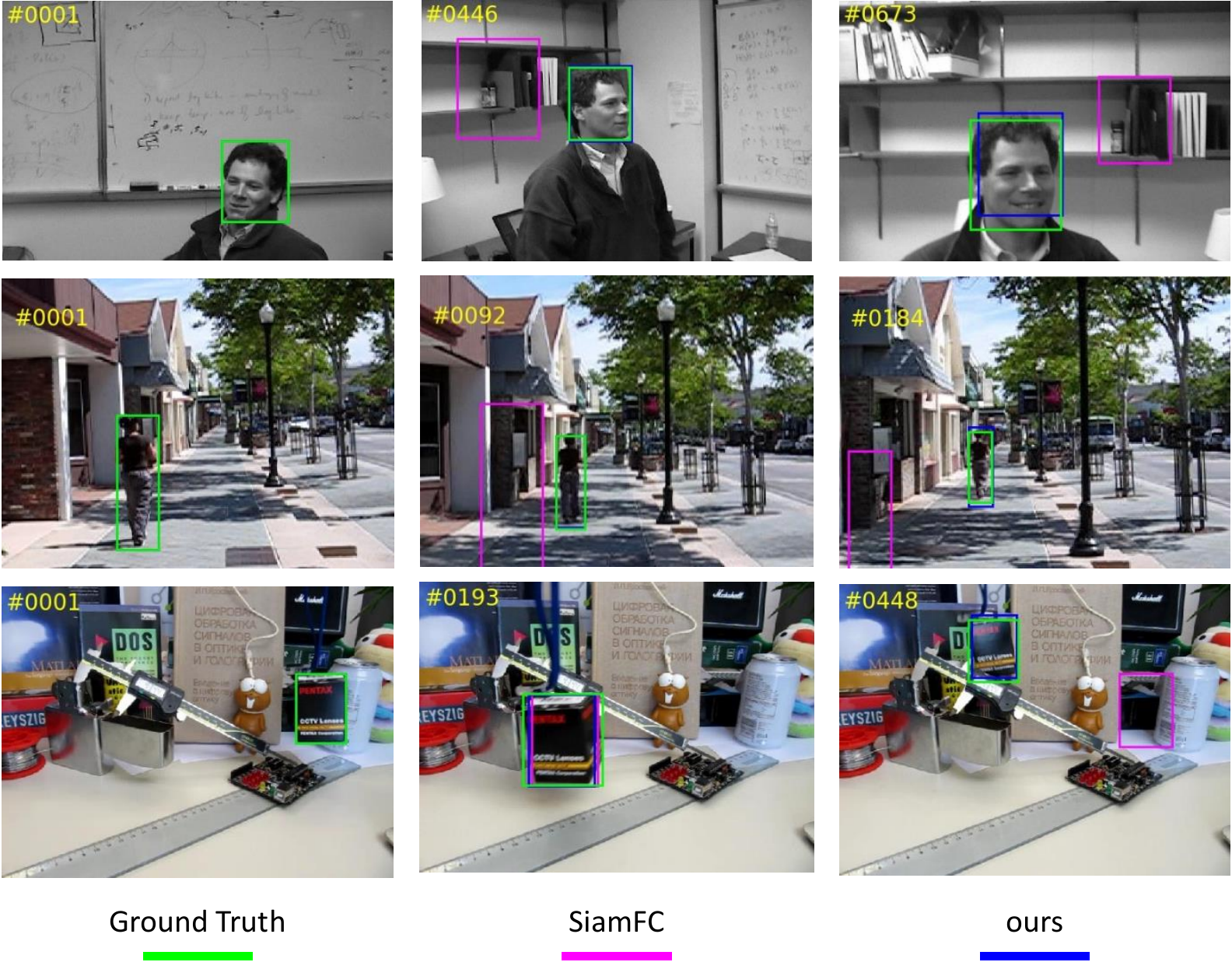} 
    \end{center}
    \caption{Comparing the tracking results of SiamFC and our tracker. Thanks to the semantic features, our tracker successfully follows the target object in case of shooting angle change or scale change, when SiamFC fails. }
    \vspace{-10pt}
    \label{fig:examples}
\end{figure}

In recent years, deep convolutional neural networks (CNNs) demonstrated their superior capabilities in various vision tasks. They have also significantly advanced the state-of-the-art of object tracking. Some trackers \cite{DeepSRDCF, HCF, HDT, CCOT, ECO} integrate deep features into conventional tracking approaches and benefit from the expressive power of CNN features. Some others \cite{MDNET, TCNN, BranchOut, STCT} directly use CNNs as classifiers and take full advantage of end-to-end training. Most of these approaches adopt online training to boost the tracking performance. However, due to the high volume of CNN features and the complexity of deep neural networks, it is computationally expensive to perform online training. As a result, most online CNN-based trackers have a far less operational speed than real-time. 


Meanwhile, there emerge two real-time CNN-based trackers \cite{GOTURN,SiamFC} which achieve high tracking speed by completely avoiding online training. While GOTURN \cite{GOTURN} treats object tracking as a box regression problem, SiamFC \cite{SiamFC} treats it as a similarity learning problem. It appears that SiamFC achieves a much better performance than GOTURN. This owes to the fully convolutional network architecture, which allows SiamFC to make full use of the offline training data and make itself highly discriminative.
However, the generalization capability remains quite poor and it encounters difficulties when the target has significant appearance change, as shown in Fig.\ref{fig:examples}. As a result, SiamFC still has a performance gap to the best online tracker. 


In this paper, we aim to improve the generalization capability of SiamFC. It is widely understood that, in a deep CNN trained for image classification task, features from deeper layers contain stronger semantic information and is more invariant to object appearance changes. These semantic features are an ideal complement to the appearance features trained in a similarity learning problem.  
Inspired by this observation, we design SA-Siam, which is a twofold Siamese network comprised of a semantic branch and an appearance branch. Each branch is a Siamese network computing the similarity scores between the target image and a search image. In order to maintain the heterogeneity of the two branches, they are separately trained and not combined until the similarity score is obtained by each branch. 

For the semantic branch, we further propose a channel attention mechanism to achieve a minimum degree of target adaptation. The motivation is that different objects activate different sets of feature channels. We shall give higher weights to channels that play more important roles in tracking specific targets. This is realized by computing channel-wise weights based on the channel responses at the target object and in the surrounding context. This simplest form of target adaptation improves the discrimination power of the tracker. Evaluations show that our tracker outperforms all other real-time trackers by a large margin on OTB-2013/50/100 benchmarks, and achieves state-of-the-art performance on VOT benchmarks. 



The rest of the paper is organized as follows. We first introduce related work in Section \ref{sec:related}. Our approach is described in Section \ref{sec:approach}. The experimental results are presented in Section \ref{sec:experiment}. Finally, Section \ref{sec:conclusion} concludes the paper. 

\begin{figure*}
    \begin{center}
    \includegraphics[width=0.95\textwidth]{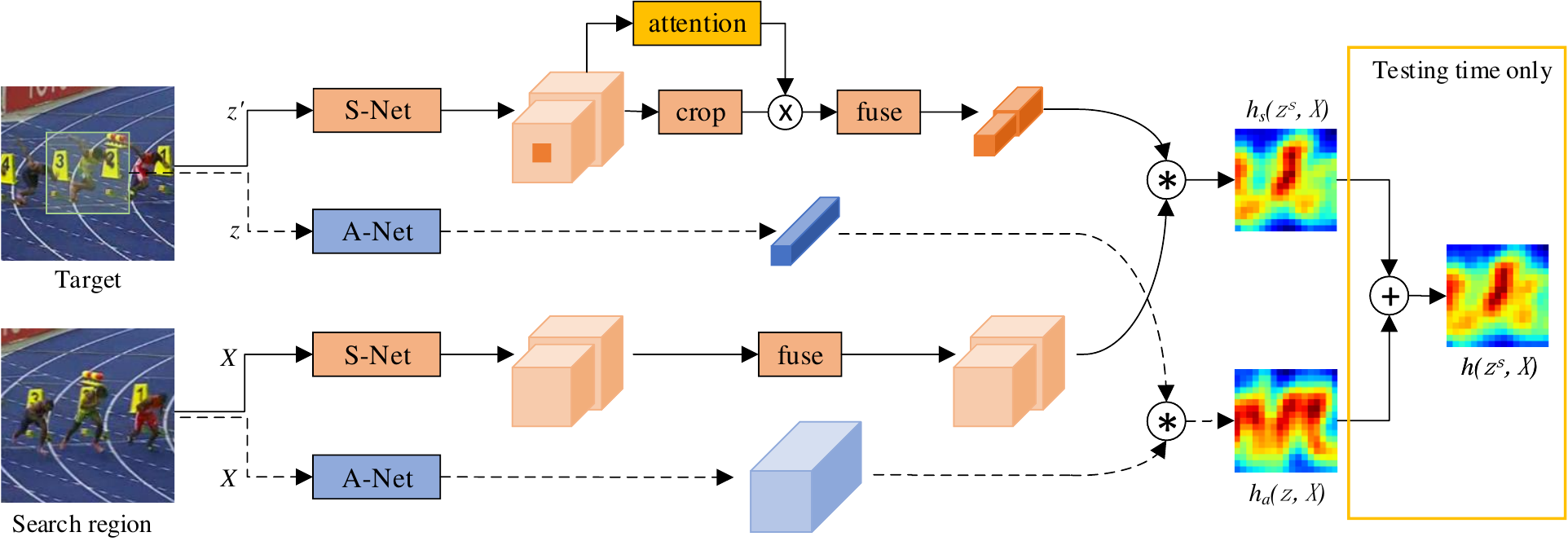} 
    \end{center}
    \caption{The architecture of the proposed twofold SA-Siam network. A-Net indicates the appearance network. The network and data structures connected with dotted lines are exactly the same as SiamFC \cite{SiamFC}. S-Net indicates the semantic network. Features from the last two convolution layers are extracted. The channel attention module determines the weight for each feature channel based on both target and context information. The appearance branch and the semantic branch are separately trained and not combined until testing time. }
    \vspace{-10pt}
    \label{fig:framework}
\end{figure*}


\section{Related Work} \label{sec:related} 
\subsection{Siamese Network Based Trackers} 

Visual object tracking can be modeled as a similarity learning problem. By comparing the target image patch with the candidate patches in a search region, we can track the object to the location where the highest similarity score is obtained. A notable advantage of this method is that it needs no or little online training. Thus, real-time tracking can be easily achieved. 

Similarity learning with deep CNNs is typically addressed using Siamese architectures \cite{SiamRep}. The pioneering work for object tracking is the fully convolutional Siamese network (SiamFC) \cite{SiamFC}. The advantage of a fully-convolutional network is that, instead of a candidate patch of the same size of the target patch, one can provide as input to the network a much larger search image and it will compute the similarity at all translated sub-windows on a dense grid in a single evaluation \cite{SiamFC}. This advantage is also reflected at training time, where every sub-window effectively represents a useful sample with little extra cost. 


There are a large number of follow-up work\cite{RFL,CFNET,EAST,SINT,DSiam} of SiamFC. EAST\cite{EAST} attempts to speed up the tracker by early stopping the feature extractor if low-level features are sufficient to track the target. CFNet\cite{CFNET} introduces correlation filters for low level CNNs features to speed up tracking without accuracy drop. 
SINT \cite{SINT} incorporates the optical flow information and achieves better performance. However, since computing optical flow is computationally expensive, SINT only operates at 4 frames per second (fps). DSiam\cite{DSiam} attempts to online update the embeddings of tracked target. Significantly better performance is achieved without much speed drop. 

SA-Siam inherits network architecture from SiamFC. We intend to improve SiamFC with an innovative way to utilize heterogeneous features.

\subsection{Ensemble Trackers} 

Our proposed SA-Siam is composed of two separately trained branches focusing on different types of CNN features. It shares some insights and design principles with ensemble trackers. 

HDT\cite{HDT} is a typical ensemble tracker. It constructs trackers using correlation filters (CFs) with CNN features from each layer, and then uses an adaptive Hedge method to hedge these CNN trackers. TCNN\cite{TCNN} maintains multiple CNNs in a tree structure to learn ensemble models and estimate target states. STCT\cite{STCT} is a sequential training method for CNNs to effectively transfer pre-trained deep features for online applications. An ensemble CNN-based classifier is trained to reduce correlation across models. BranchOut\cite{BranchOut} employs a CNN for target representation, which has a common convolutional layers but has multiple branches of fully connected layers. 
It allows each branch to have a different number of layers so as to maintain variable abstraction levels of target appearances. PTAV\cite{PTAV} keeps two classifiers, one acting as the tracker and the other acting as the verifier. The combination of an efficient tracker which runs for sequential frames and a powerful verifier which only runs when necessary strikes a good balance between speed and accuracy. 

A common insight of these ensemble trackers is that it is possible to make a strong tracker by utilizing different layers of CNN features. Besides, the correlation across models should be weak. In SA-Siam design, the appearance branch and the semantic branch use features at very different abstraction levels. Besides, they are not jointly trained to avoid becoming homogeneous.

\subsection{Adaptive Feature Selection}
Different features have different impacts on different tracked target. Using all features available for a single object tracking is neither efficient nor effective. In SCT \cite{SCT} and ACFN \cite{ACFN}, the authors build an attention network to select the best module among several feature extractors for the tracked target. HART \cite{HART} and RATM\cite{RATM} use RNN with attention model to separate where and what processing pathways to actively suppress irrelevant visual features. Recently, SENet\cite{SENET} demonstrates the effectiveness of channel-wise attention on image recognition tasks. 

In our SA-Siam network, we perform channel-wise attention based on the channel activations. It can be looked on as a type of target adaptation, which potentially improves the tracking performance.



\section{Our Approach} \label{sec:approach}

We propose a twofold fully-convolutional siamese network for real-time object tracking. The fundamental idea behind this design is that the appearance features trained in a similarity learning problem and the semantic features trained in an image classification problem complement each other, and therefore should be jointly considered for robust visual tracking. 


\subsection{SA-Siam Network Architecture}
The network architecture of the proposed SA-Siam network is depicted in Fig.~\ref{fig:framework}. The input of the network is a pair of image patches cropped from the first (target) frame of the video sequence and the current frame for tracking. We use notations $z$, $z^s$ and $X$ to denote the images of target, target with surrounding context and search region, respectively. Both $z^s$ and $X$ have a size of $W_s \times H_s \times 3$. The exact target $z$ has a size of $W_t \times H_t \times 3$ ($W_t < W_s$ and $H_t < H_s$), locates in the center of $z^s$. $X$ can be looked on as a collection of candidate image patches $x$ in the search region which have the same dimension as $z$. 

SA-Siam is composed of the appearance branch (indicated by blue blocks in the figure) and the semantic branch (indicated by orange blocks). 
The output of each branch is a response map indicating the similarity between target $z$ and candidate patch $x$ within the search region $X$. The two branches are separately trained and not combined until testing time. 

\textbf{The appearance branch:} The appearance branch takes $(z,X)$ as input. It clones the SiamFC network \cite{SiamFC}. The convolutional network used to extract appearance features is called A-Net, and the features extracted are denoted by $f_a(\cdot)$. The response map from the appearance branch can be written as:
\begin{equation}
h_a(z, X) = corr(f_a(z), f_a(X)),
\end{equation}
where $corr(\cdot)$ is the correlation operation. All the parameters in the A-Net are trained from scratch in a similarity learning problem. In particular, with  abundant pairs $(z_i, X_i)$ from training videos and corresponding ground-truth response map $Y_i$ of the search region, A-Net is optimized by minimizing the logistic loss function $L(\cdot)$ as follows:
\begin{equation}
\arg \min_{\theta_a} \frac{1}{N} \sum_{i=1}^N{\left\lbrace L \left( h_a(z_i, X_i{;}\; \theta_a), Y_i \right) \right\rbrace},
\end{equation} 
where $\theta_a$ denotes the parameters in A-Net, $N$ is the number of training samples.

\textbf{The semantic branch:} The semantic branch takes $(z^s, X)$ as input. We directly use a CNN pretrained in the image classification task as S-Net and fix all its parameters during training and testing. We let S-Net to output features from the last two convolutional layers ($conv4$ and $conv5$ in our implementation), since they provide different levels of abstraction. The low-level features are not extracted. 

Features from different convolutional layers have different spatial resolution. For simplicity of notation, we denote the concatenated multilevel features by $f_s(\cdot)$. In order to make the semantic features suitable for the correlation operation, we insert a fusion module, implemented by $1 \times 1$ ConvNet, after feature extraction. The fusion is performed within features of the same layer. The feature vector for search region $X$ after fusion can be written as $g(f_s(X))$.

The target processing is slightly different. S-Net takes $z^s$ as the target input. $z^s$ has $z$ in its center and contains the context information of the target. Since S-Net is fully convolutional, we can easily obtain $f_s(z)$ from $f_s(z^s)$ by a simple crop operation. The attention module takes $f_s(z^s)$ as input and outputs the channel weights $\xi$. The details of the attention module is included in the next subsection. The features are multiplied by channel weights before they are fused by $1 \times 1$ ConvNet. As such, the response map from the semantic branch can be written as:
\begin{equation}
h_s(z^s, X) = corr\left(
g\left(\xi \cdot f_s(z)\right), 
g\left(f_s(X)\right)
\right),
\end{equation}
where $\xi$ has the same dimension as the number of channels in $f_s(z)$ and $\cdot$ is element-wise multiplication. 

In the semantic branch, we only train the fusion module and the channel attention module. With training pairs $(z_i^s, X_i)$ and ground-truth response map $Y_i$, the semantic branch is optimized by minimizing the following logistic loss function $L(\cdot)$:
\begin{equation}
\arg \min_{\theta_s} \frac{1}{N} \sum_{i=1}^N{\left\lbrace L \left( h_s(z_i^s, X_i{;}\; \theta_s), Y_i \right) \right\rbrace},
\end{equation} 
where $\theta_s$ denotes the trainable parameters, $N$ is the number of training samples.

During testing time, the overall heat map is computed as the weighted average of the heat maps from the two branches: 
\begin{equation}
h(z^s,X) = \lambda h_a(z,X) + (1-\lambda)h_s(z^s,X),
\end{equation}
where $\lambda$ is the weighting parameter to balance the importance of the two branches. In practice, $\lambda$ can be estimated from a validation set. The position of the largest value in $h(z^s, X)$ suggests the center location of the tracked target. Similar to SiamFC \cite{SiamFC}, we use multi-scale inputs to deal with scale changes. We find that using three scales strikes a good balance between performance and speed. 

\subsection{Channel Attention in Semantic Branch}

High-level semantic features are robust to appearance changes of objects, and therefore make a tracker more generalized but less discriminative. In order to enhance the discriminative power of the semantic branch, we design a channel attention module. 


Intuitively, different channels play different roles in tracking different targets. Some channels may be extremely important in tracking certain targets while being dispensable in tracking others. If we could adapt the channel importance to the tracking target, we achieve the minimum functionality of target adaptation. In order to do so, not only the target is relevant, the surrounding context of the target also matters. Therefore, in our proposed attention module, the input is the feature map of $z^s$ instead of $z$.
 
\begin{figure}[t!]
    \begin{center}
    \includegraphics[width=\columnwidth]{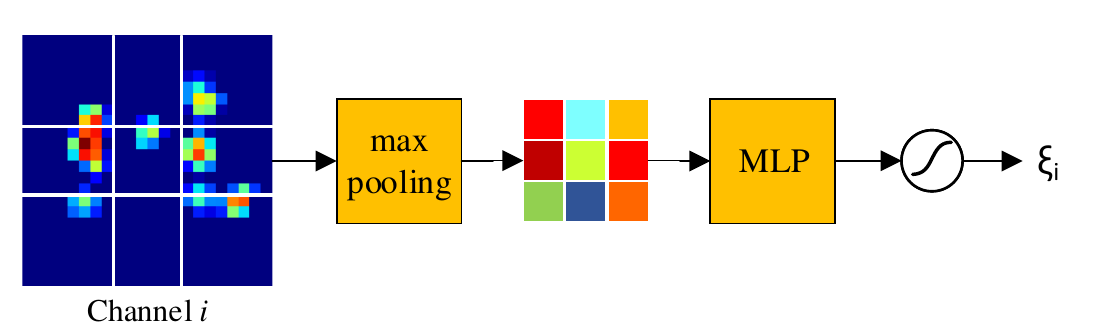} 
    \end{center}
    \caption{Channel-wise attention generates the weighing coefficient $\xi_i$ for channel $i$ through max pooling and multilayer perceptron (MLP).} 
    \vspace{-10pt}
    \label{fig:attention}
\end{figure}

The attention module is realized by channel-wise operations. Fig.~\ref{fig:attention} shows the attention process for the $i^{th}$ channel. Take a $conv5$ feature map as an example, the spatial dimension is $22 \times 22$ in our implementation. We divide the feature map into $3 \times 3$ grids, and the center grid with size $6 \times 6$ corresponds to the tracking target $z$. Max pooling is performed within each grid, and then a two-layer multilayer perceptron (MLP) is used to produce a coefficient for this channel. Finally, a Sigmoid function with bias is used to generate the final output weight $\xi_i$. The MLP module shares weights across channels extracted from the same convolutional layer. 

Note that attention is only added to the target processing. All the activations in channel $i$ for the target patch are multiplied by $\xi_i$. 
Therefore, this module is passed only once for the first frame of a tracking sequence. The computational overhead is negligible. 

\subsection{Discussions of Design Choices}
\textbf{We separately train the two branches.} We made this choice based on the following observations. For some training samples, tracking with semantic clues may be much easier than with appearance clues, and for some others, it could just be the opposite. Let us take the former case as an example. If the two branches are jointly trained, the overall loss could be small when the semantic branch has a discriminative heat map and the appearance branch has a non-informative heat map. Then, these training samples do not play their part in optimizing the appearance branch. As a result, both branches are less powerful when they are jointly trained than separately trained. 


\textbf{We do not fine-tune S-Net.} A common practice in transfer learning is to fine-tune the pre-trained network in the new problem. We choose not to do so because fine-tuning S-Net using the same procedure as we train A-Net will make the two branches homogeneous. Although fine-tuning S-Net may improve the tracking performance of the semantic branch alone, the overall performance could be unfavorably impacted. We have carried out experiments to verify this design choice, although we do not include them in this paper due to space limitation. 

\textbf{We keep A-Net as it is in SiamFC.} Using multilevel features and adding channel attention significantly improve the semantic branch, but we do not apply them to the appearance branch. This is because appearance features from different convolutional layers do not have significant difference in terms of expressiveness. We cannot apply the same attention module to the appearance branch because high-level semantic features are very sparse while appearance features are quite dense. A simple max pooling operation could generate a descriptive summary for semantic features but not for appearance features. Therefore, a much more complicated attention module would be needed for A-Net, and the gain may not worth the cost. 

\section{Experiments} \label{sec:experiment}


\subsection{Implementation Details}

\textbf{Network structure:} Both A-Net and S-Net use AlexNet\cite{ALEXNET}-like network as base network. The A-Net has exactly the same structure as the SiamFC network \cite{SiamFC}. The S-Net is loaded from a pretrained AlexNet on ImageNet\cite{ILSVRC}. A small modification is made to the stride so that the last layer output from S-Net has the same dimension as A-Net. 

In the attention module, the pooled features of each channel are stacked into a 9-dimensional vector. The following MLP has $1$ hidden layer with nine neurons. The non-linear function of the hidden layer is ReLU. The MLP is followed by a Sigmoid function with bias $0.5$. This is to ensure that no channel will be suppressed to zero. 

\textbf{Data Dimensions:} In our implementation, the target image patch $z$ has a dimension of $127 \times
127 \times 3$, and both $z^s$ and $X$ have a dimension of $255 \times 255 \times 3$. The output features of A-Net for $z$ and $X$ have a dimension of $6 \times 6 \times 256$ and $22 \times 22 \times 256$, respectively. The $conv4$ and $conv5$ features from S-Net have dimensions of $24 \times 24 \times 384$ and $22 \times 22 \times 256$ channels for $z^s$ and $X$. The $1 \times 1$ ConvNet for these two sets of features outputs 128 channels each (which adds up to 256), with the spatial resolution unchanged. The response maps have the same dimension of $17 \times 17$. 

\textbf{Training:} Our approach is trained offline on the ILSVRC-2015~\cite{ILSVRC} video dataset and we only use color images. Among a total of more than 4,000 sequences, there are around 1.3 million frames and about 2 million tracked objects with ground truth bounding boxes. For each tracklet, we randomly pick a pair of images and crop $z^s$ from one image with $z$ in the center and crop $X$ from the other image with the ground truth target in the center. 
Both branches are trained for $30$ epochs with learning rate $0.01$ in the first $25$ epochs and learning rate $0.001$ in the last five epochs.

We implement our model in TensorFlow\cite{Tensorflow} 1.2.1 framework. Our experiments are performed on a PC with a Xeon E5 2.4GHz CPU and a GeForce GTX Titan X GPU. The average testing speed of SA-Siam is 50 fps. 

\textbf{Hyperparameters:} 
The two branches are combined by a weight $\lambda$. This hyperparameter is estimated on a small validation set TC128\cite{TC128} excluding sequences in OTB benchmarks. We perform a grid search from $0.1$ to $0.9$ with step $0.2$. Evaluations suggest that the best performance is achieved when $\lambda=0.3$. We use this value for all the test sequences. During evaluation and testing, three scales are searched to handle scale variations.

\begin{table*}
    \begin{center}
    \small{
    \begin{tabular}{cccc|cc|cc|cc}
        \hline
         \multicolumn{4}{c}{} & \multicolumn{2}{|c}{OTB-2013} & \multicolumn{2}{|c}{OTB-50} & \multicolumn{2}{|c}{OTB-100}\\
         App. & Sem. & ML & Att. & AUC & Prec. &  AUC & Prec.  &  AUC & Prec. \\
         \hline
         \checkmark & &&                               & 0.599 & 0.811 & 0.520 & 0.718 & 0.585 & 0.790\\
         &\checkmark &&                                & 0.607 & 0.821 & 0.517 & 0.715 & 0.588 & 0.791\\
         \checkmark&\checkmark &&                      & 0.649 & 0.862 & 0.583 & 0.792 & 0.637 & 0.843\\
         \checkmark&\checkmark & \checkmark &          & 0.656 & 0.865 & 0.581 & 0.778 & 0.641 & 0.841\\
         \checkmark&\checkmark & &\checkmark           & 0.650 & 0.861 & 0.591 & 0.803 & 0.642 & 0.849\\
         \checkmark&\checkmark & \checkmark &\checkmark& 0.676 & 0.894 & 0.610 & 0.823 & 0.656 & 0.864\\
         \hline
    \end{tabular}}
    \end{center}
    \caption{Ablation study of SA-Siam on OTB benchmarks. App. and Sem. denote appearance model and semantic model. ML means using multilevel feature and Att. denotes attention module.}
    \vspace{-10pt}
    \label{tab:res-emp}
\end{table*}

\subsection{Datasets and Evaluation Metrics}

\textbf{OTB:} The object tracking benchmarks (OTB)\cite{OTB13, OTB15} consist of three datasets, namely OTB-2013, OTB-50 and OTB-100. They have 51, 50 and 100 real-world target for tracking, respectively. There are eleven interference attributes to which all sequences belong. 

The two standard evaluation metrics on OTB are success rate and precision. For each frame, we compute the IoU (intersection over union) between the tracked and groundtruth bounding boxes, as well as the distance of their center locations. A success plot can be obtained by evaluating the success rate at different IoU thresholds. Conventionally, the area-under-curve (AUC) of the success plot is reported. The precision plot can be acquired in a similar way, but usually the representative precision at the threshold of 20 pixels is reported. We use the standard OTB toolkit to obtain all the numbers. 


\textbf{VOT:} The visual object tracking (VOT) benchmark has many versions, and we use the most recent ones:  VOT2015\cite{VOT2015}, VOT2016\cite{VOT2016} and VOT2017\cite{VOT2017}. VOT2015 and VOT2016 contain the same sequences, but the ground truth label in VOT2016 is more accurate than that in VOT2015. 
In VOT2017, ten sequences from VOT2016 are replaced by new ones.

The VOT benchmarks evaluate a tracker by applying a reset-based methodology. Whenever a tracker has no overlap with the ground truth, the tracker will be re-initialized after five frames. Major evaluation metrics of VOT benchmarks are accuracy (A), robustness (R) and expected average overlap (EAO). A good tracker has high A and EAO scores but low R scores.


\subsection{Ablation Analysis}

We use experiments to verify our claims and justify our design choices in SA-Siam. We use the OTB benchmark for the ablation analysis. 

\textbf{The semantic branch and the appearance branch complement each other.} 
We evaluate the performances of the semantic model alone (model $S_1$) and the appearance model alone (model $A_1$). $S_1$ is a basic version with only S-Net and fusion module. Both $S_1$ and $A_1$ are trained from scratch with random initialization. The results are reported in the first two rows in Table \ref{tab:res-emp}. The third row presents the results of an SA-Siam model which combines $S_1$ and $A_1$. The improvement is huge and the combined model achieves state-of-the-art performance even without multilevel semantic features or channel attention.

\begin{table}
    \begin{center}
    \small{
    \begin{tabular}{l|c c c c}
        \hline
         Model   & $S_2$ & $A_2$ & $A_1A_2$ & $S_1S_2$ \\
         \hline
         AUC     & 0.606 & 0.602 & 0.608    & 0.602     \\
         Prec.   & 0.822 & 0.806 & 0.813    & 0.811     \\
         \hline
    \end{tabular}}
    \end{center}
    \caption{Evaluation of separate and ensemble models on OTB-2013. $A_1A_2$ is an ensemble of appearance models and $S_1S_2$ is an ensemble of basic semantic models.}
    \vspace{0pt}
    \label{tab:sacombine}
\end{table}

In order to show the advantage of using heterogeneous features, we compare SA-Siam with two simple ensemble models. We train another semantic model $S_2$ using different initialization, and take the appearance model $A_2$ published by the SiamFC authors. Then we ensemble $A_1A_2$ and $S_1S_2$. Table \ref{tab:sacombine} shows the performance of the separate and the ensemble models. It is clear that the ensemble models $A_1A_2$ and $S_1S_2$ do not perform as well as the SA-Siam model. This confirms the importance of complementary features in designing a twofold Siamese network. 


\textbf{Using multilevel features and channel attention bring gain.}
The last three rows in Table \ref{tab:res-emp} show how each component improves the tracking performance. Directly using multilevel features is slightly helpful, but there lacks a mechanism to balance the features of different abstraction levels. We find that the attention module plays an important role. It effectively balances the intra-layer and inter-layer channel importance and achieves significant gain. 

\begin{figure}
    \begin{center}
    \includegraphics[width=0.45\columnwidth]{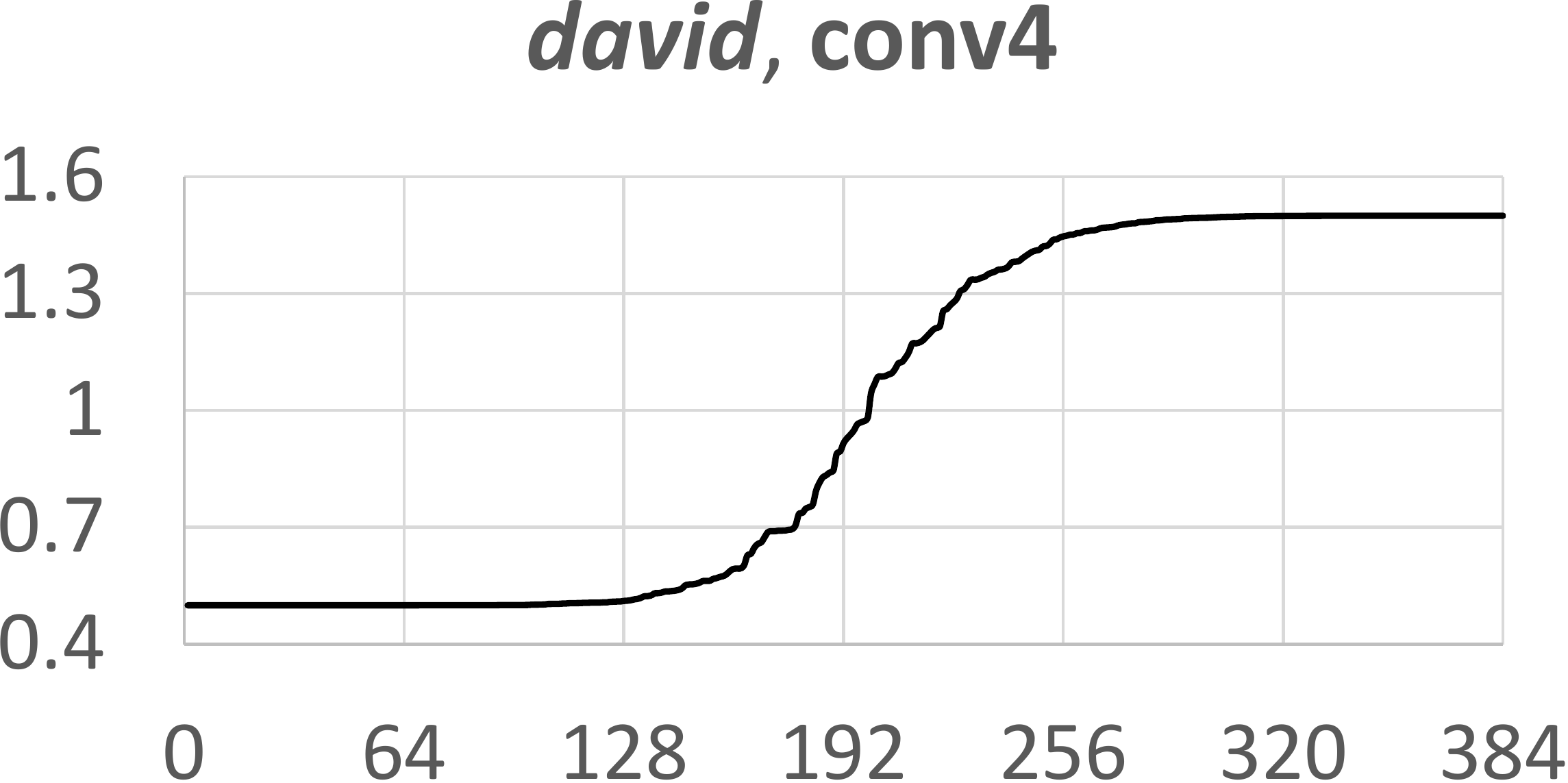} 
    \includegraphics[width=0.45\columnwidth]{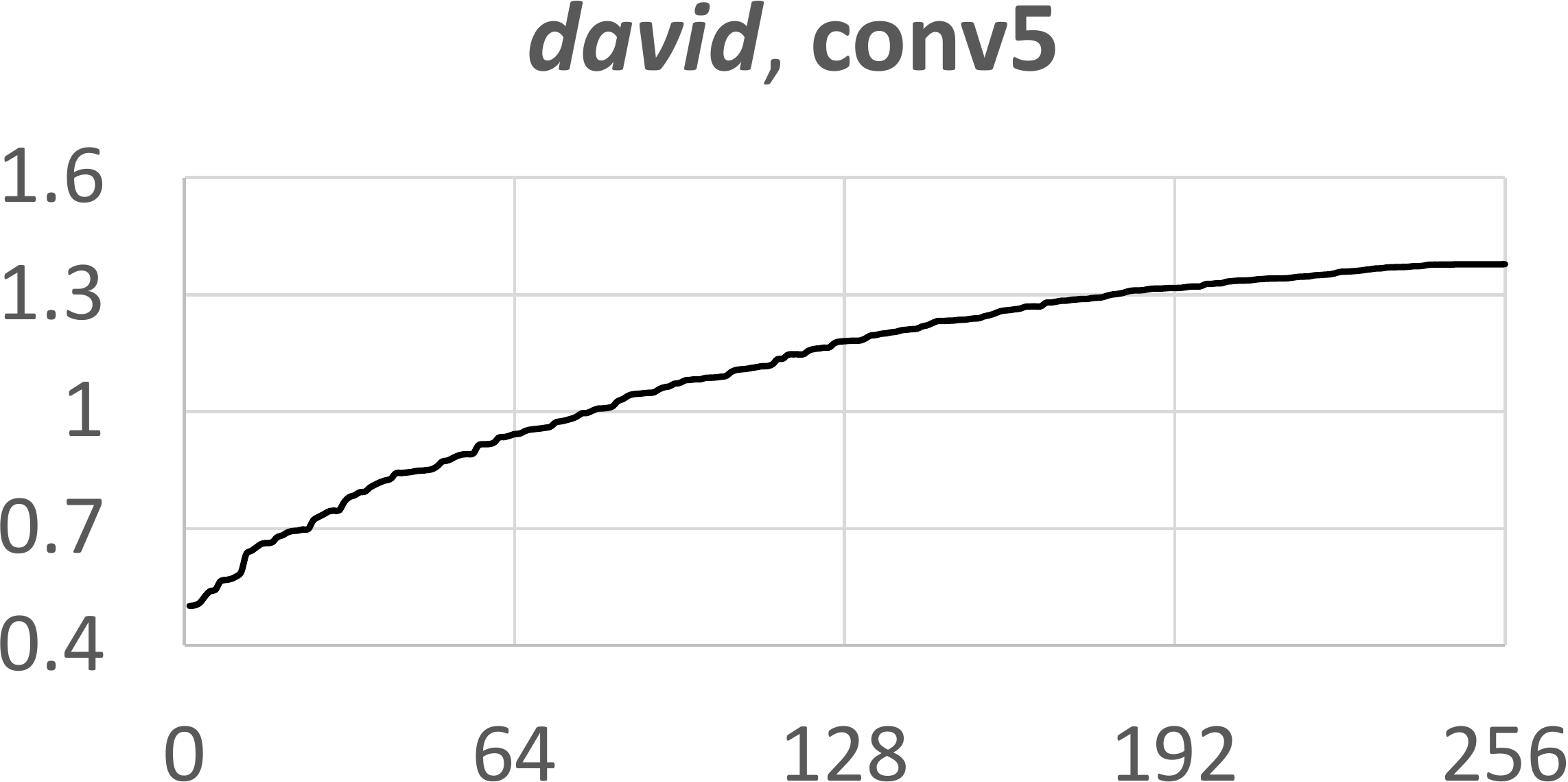} 
    \includegraphics[width=0.45\columnwidth]{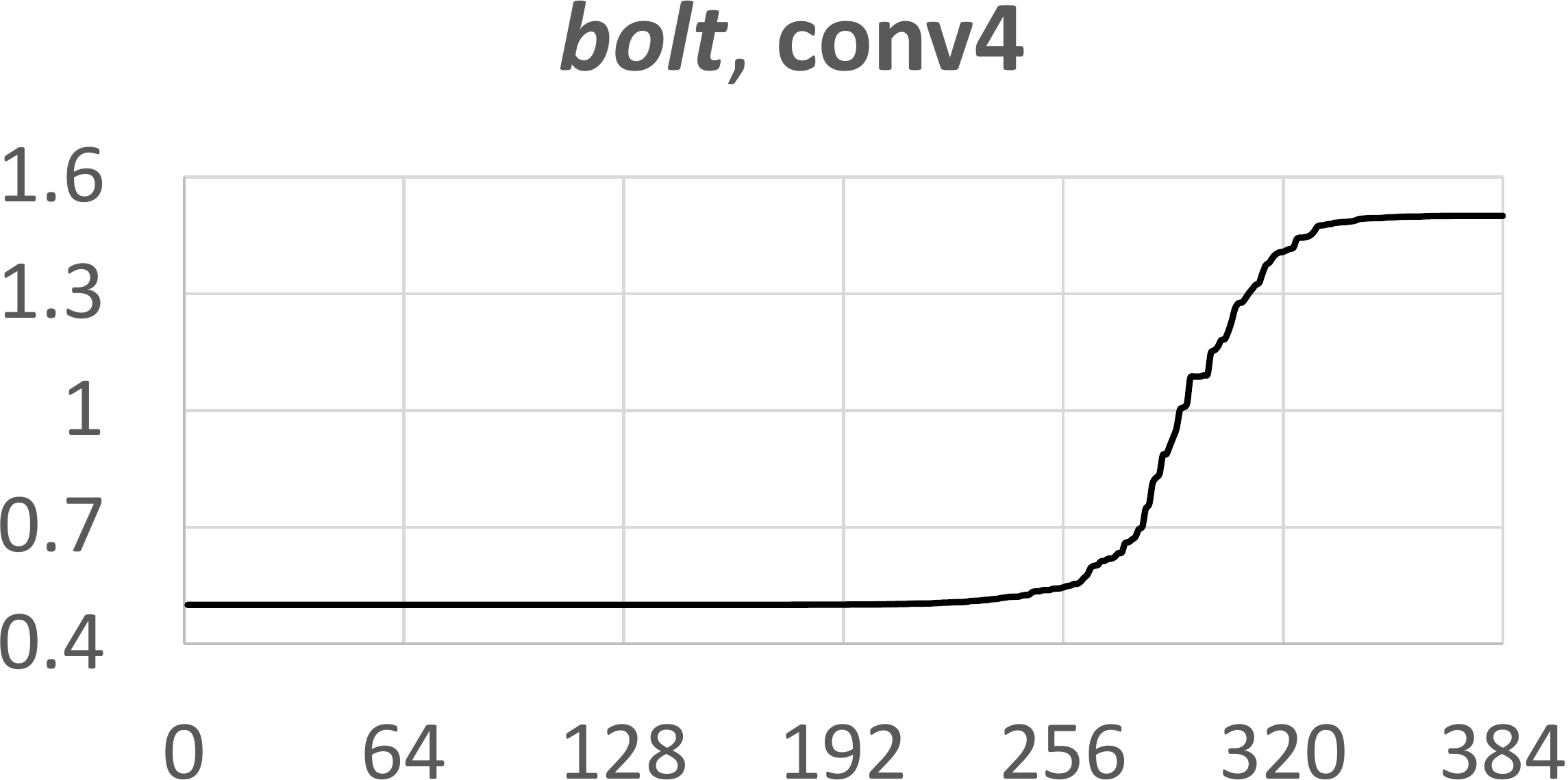}  
    \includegraphics[width=0.45\columnwidth]{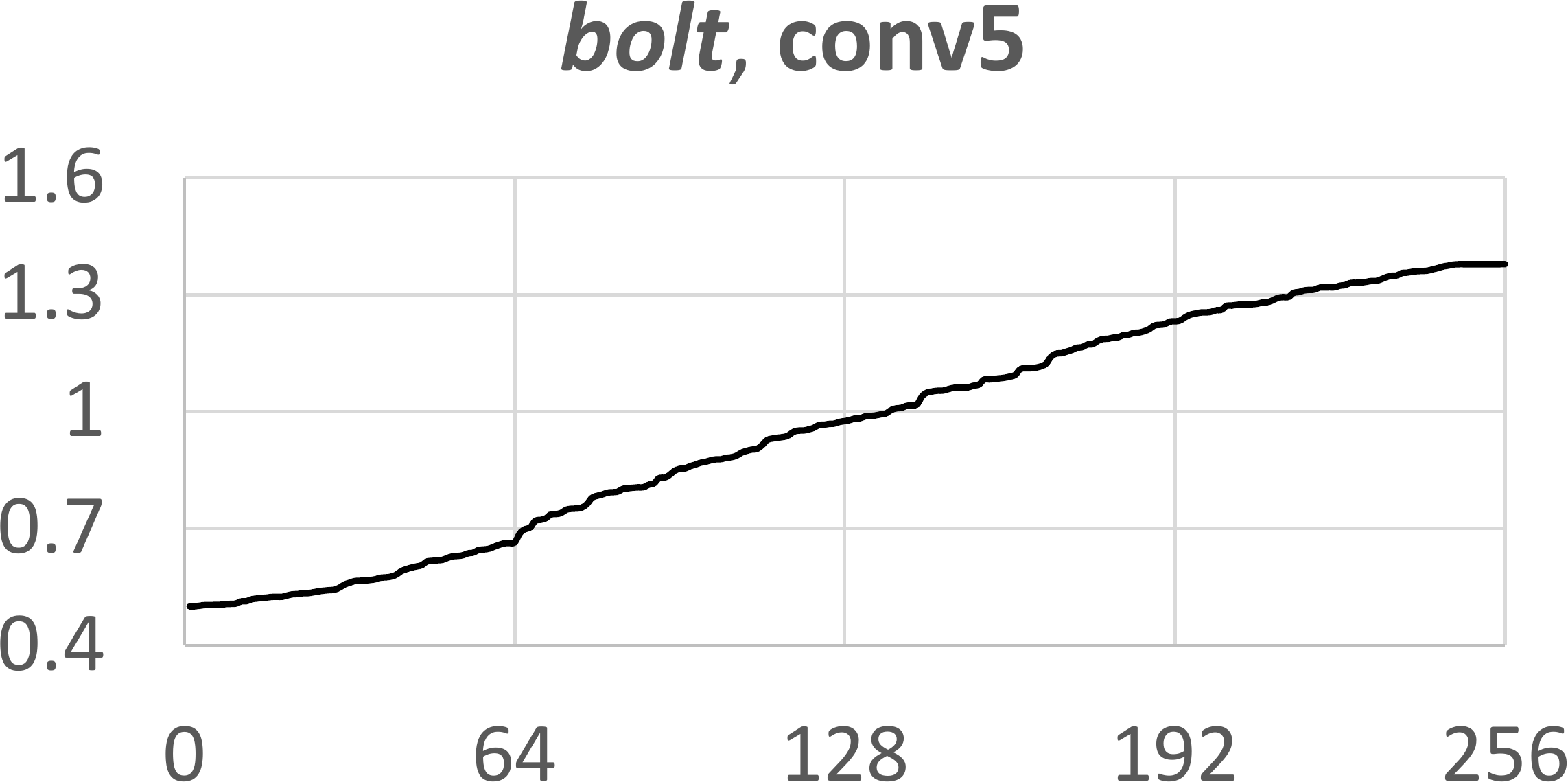} 
    \end{center}
    \caption{Visualizing the channel weights output by attention module for video \emph{david} and \emph{bolt}. Channels are sorted according to the weights. There is no correspondence between channel numbers for the two videos. }
    \vspace{-10pt}
    \label{fig:att}
\end{figure}

\begin{table*}
    \begin{center}
    \small{
    \begin{tabular}{l|l|ccccccccccc}
        \hline
         & Tracker & SA-Siam & BACF  & PTAV & ECOhc & DSiamM & EAST  & Staple   & SiamFC & CFNet & LMCF & LCT\\
         && (ours)  & \cite{BACF} &\cite{PTAV}  & \cite{ECO} &\cite{DSiam} &\cite{EAST} &\cite{STAPLE}&\cite{SiamFC} &\cite{CFNET} &\cite{LMCF} &\cite{LCT}\\ 
         \hline
         \multirow{2}{*}{OTB-2013} 
         & AUC     & \textbf{0.677}   & 0.656* & 0.663 & 0.652 & 0.656 & 0.638 & 0.593    & 0.607  & 0.611 & 0.628& 0.628 \\
         &Prec.    & \textbf{0.896}   & 0.859     & 0.895 & 0.874 & 0.891 & -     & 0.782    & 0.809  & 0.807 & 0.842& 0.848 \\
         \hline
         \multirow{2}{*}{OTB-50}
         & AUC     & \textbf{0.610}   & 0.570     & 0.581   & 0.592  & -     & -     & 0.507        & 0.516  & 0.530 & 0.533    & 0.492 \\
         &Prec.    & \textbf{0.823}   & 0.768     & 0.806    & 0.814 & -     & -     & 0.684        & 0.692  & 0.702 & 0.730    & 0.691 \\
         \hline
         \multirow{2}{*}{OTB-100} 
         & AUC     & \textbf{0.657}   & 0.621* & 0.635 &0.643 & -     & 0.629 & 0.578    & 0.582  & 0.568 & 0.580& 0.562 \\
         &Prec.    & \textbf{0.865}   & 0.822     & 0.849 & 0.856 & -     & -     & 0.784    & 0.771  & 0.748 & 0.789    & 0.762 \\
         \hline
         & FPS     & 50      & 35    & 25  & 60  & 25    & 159   & 80       & 86     & 75    & 85   & 27    \\
         \hline
    \end{tabular}}
    \end{center}
    \caption{Comparison of state-of-the-art real-time trackers on OTB benchmarks. * The reported AUC scores of BACF on OTB-2013 (excluding \emph{jogging-2}) and OTB-100 are 0.678 and 0.630, respectively. They have used a slightly different measure.}
    \vspace{-10pt}
    \label{tab:res-soa}
\end{table*}

Fig.\ref{fig:att} visualizes the channel weights for the two convolutional layers of video sequence \emph{david} and \emph{bolt}. We have used a Sigmoid function with bias $0.5$ so that the weights are in range $[0.5,1.5]$. First, we observe that the weight distributions are quite different for $conv4$ and $conv5$. This provides hints why the attention module has a larger impact on models using multilevel features. Second, the weight distributions for $conv4$ are quite different for the two videos. The attention module tends to suppress more channels from $conv4$ for video \emph{bolt}.

\textbf{Separate vs. joint training.} We have claimed that the two branches in SA-Siam should be separately trained. In order to support this statement, we tried to jointly train the two branches (without multilevel features or attention). As we anticipated, the performance is not as good as the separate-training model. The AUC and precision of the joint training model on OTB-2013/50/100 benchmarks are (0.630, 0.831), (0.546,0.739), (0.620, 0.819), respectively. 

\begin{figure}
    \begin{center}
    \includegraphics[width=0.48\columnwidth]{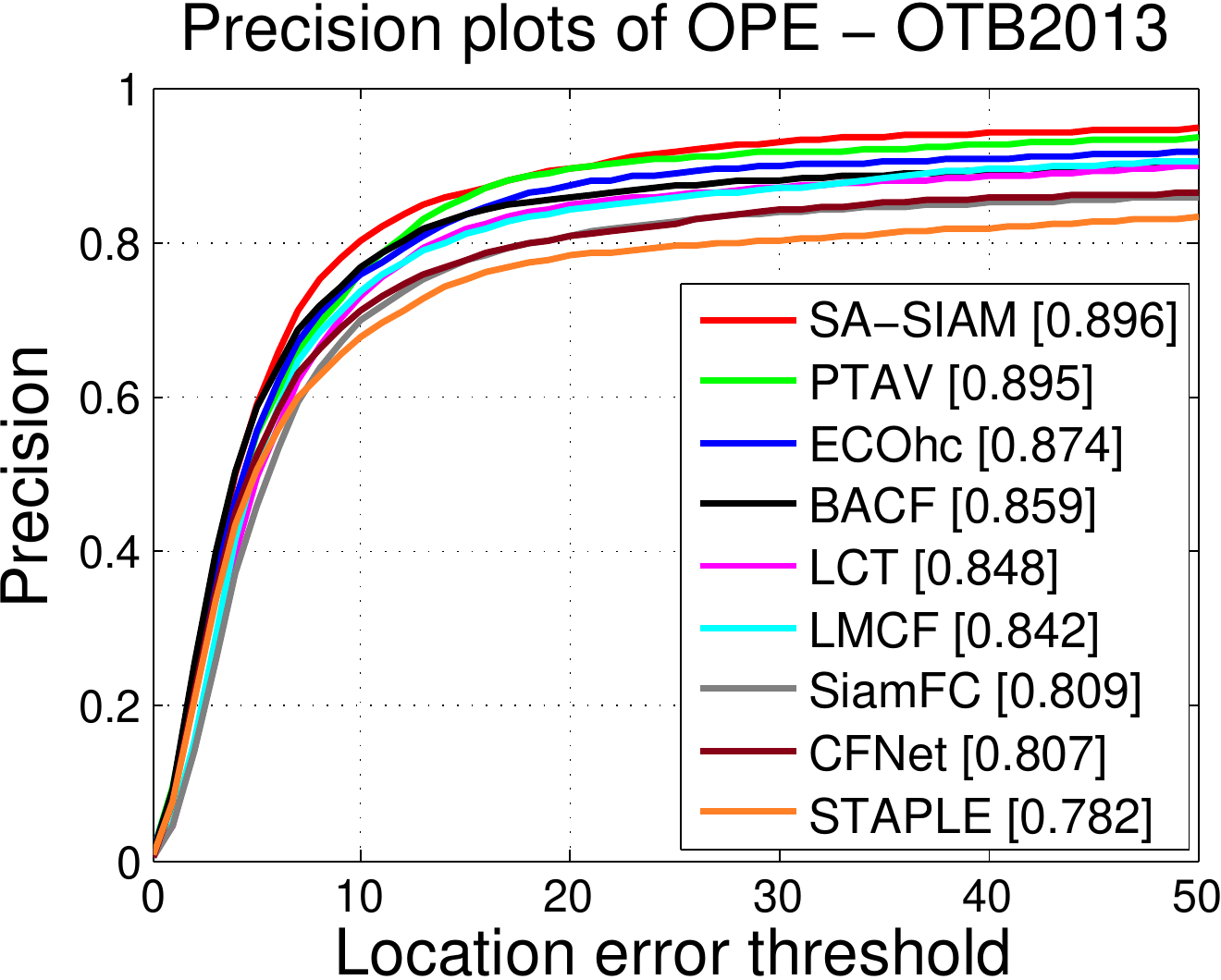} 
    \includegraphics[width=0.48\columnwidth]{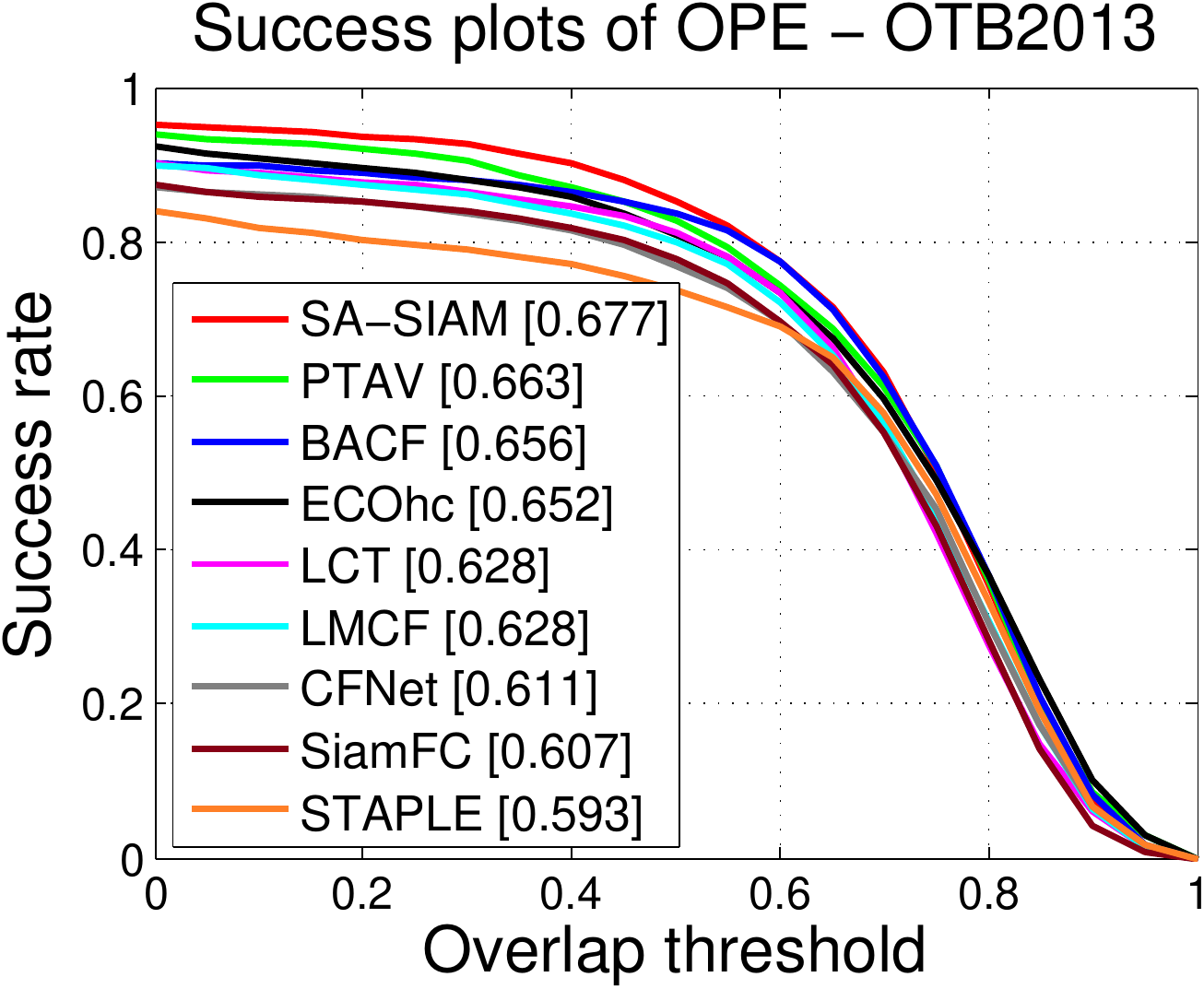} 

    \includegraphics[width=0.48\columnwidth]{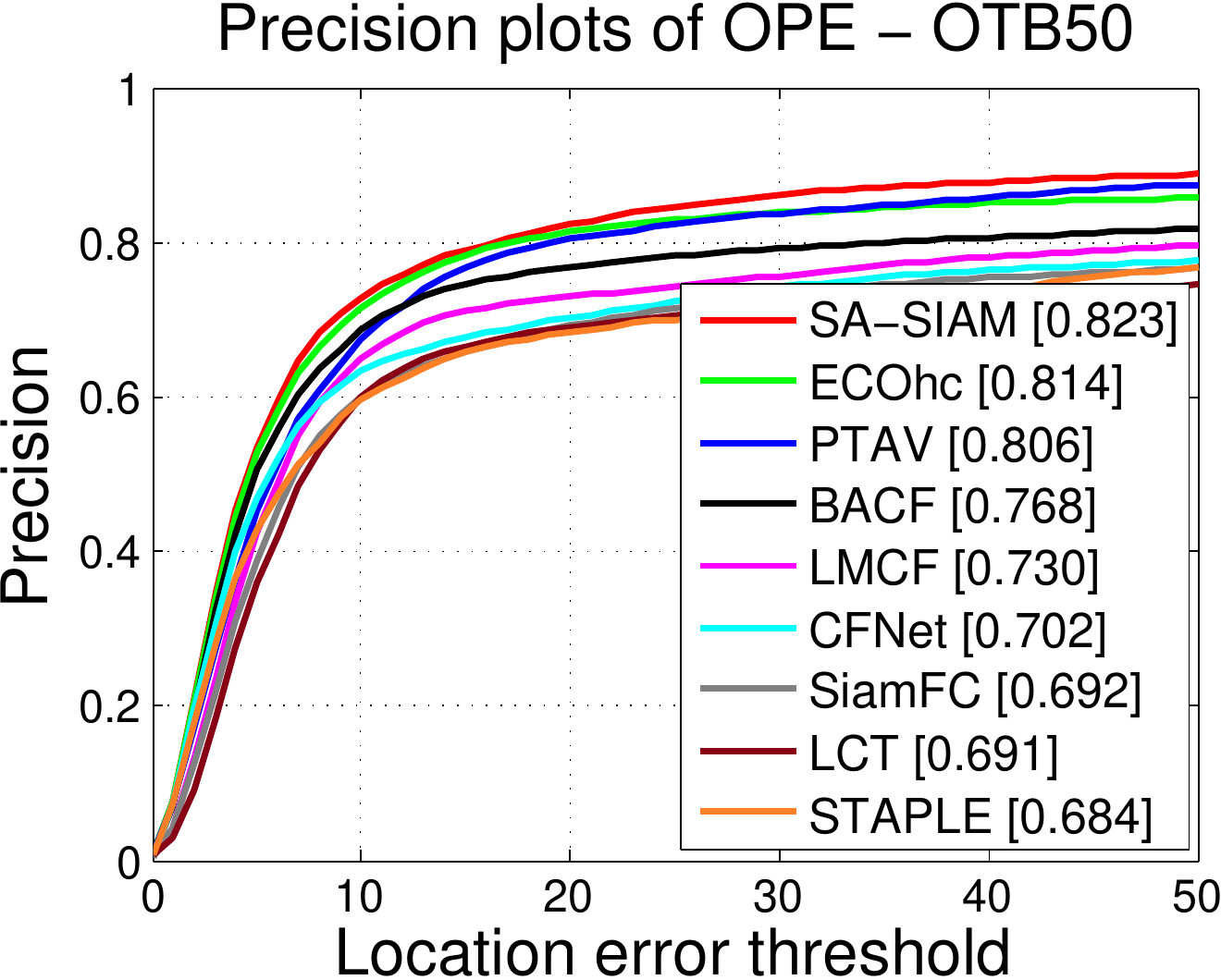} 
    \includegraphics[width=0.48\columnwidth]{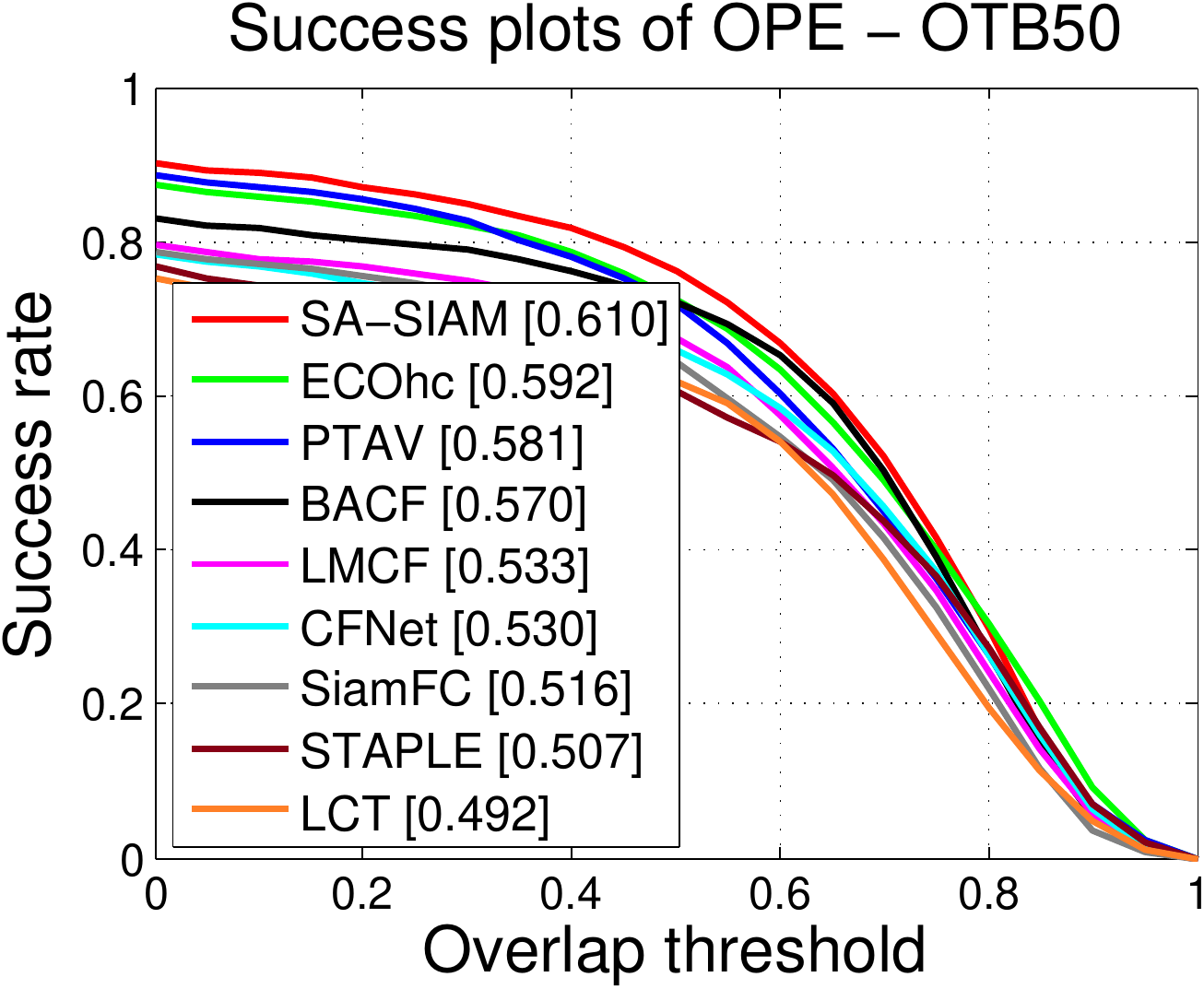} 

    \includegraphics[width=0.48\columnwidth]{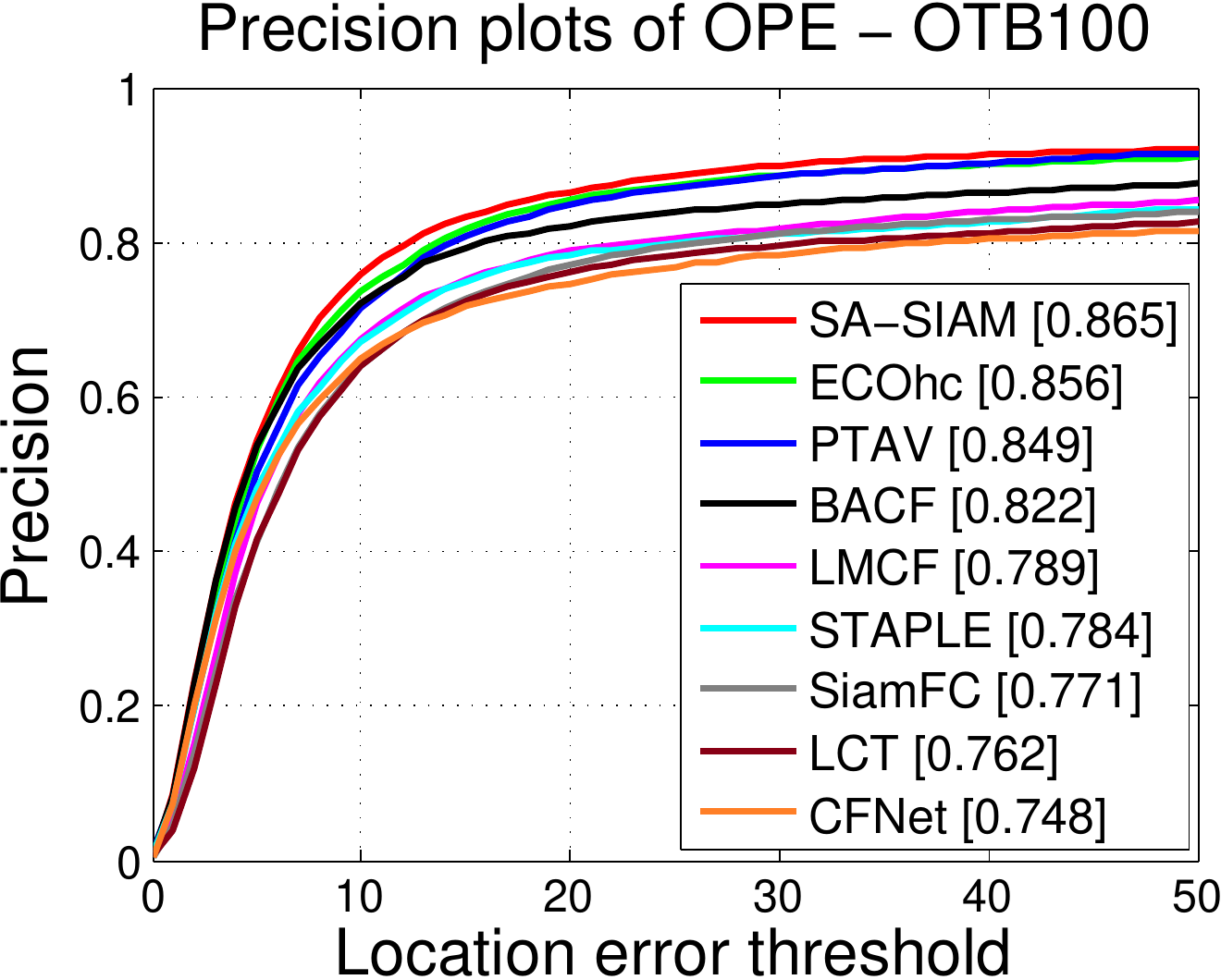} 
    \includegraphics[width=0.48\columnwidth]{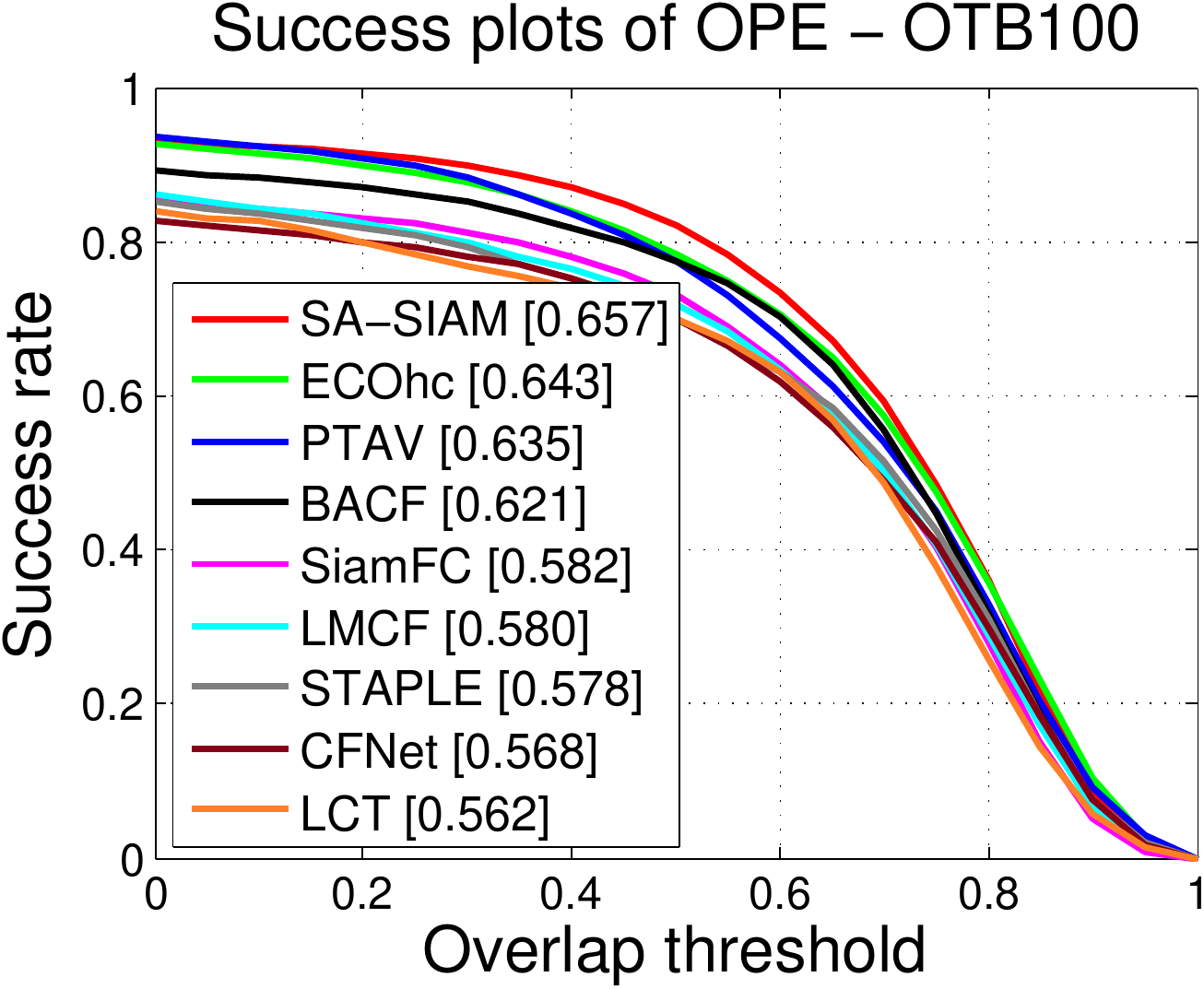} 
    \end{center}
    \caption{The precision plots and success plots over three OTB benchmarks. Curves and numbers are generated with OTB toolkit.}
    \vspace{-10pt}
    \label{fig:OPEplots}
\end{figure}

\begin{figure*}
    \begin{center}
    \includegraphics[width=0.9\textwidth]{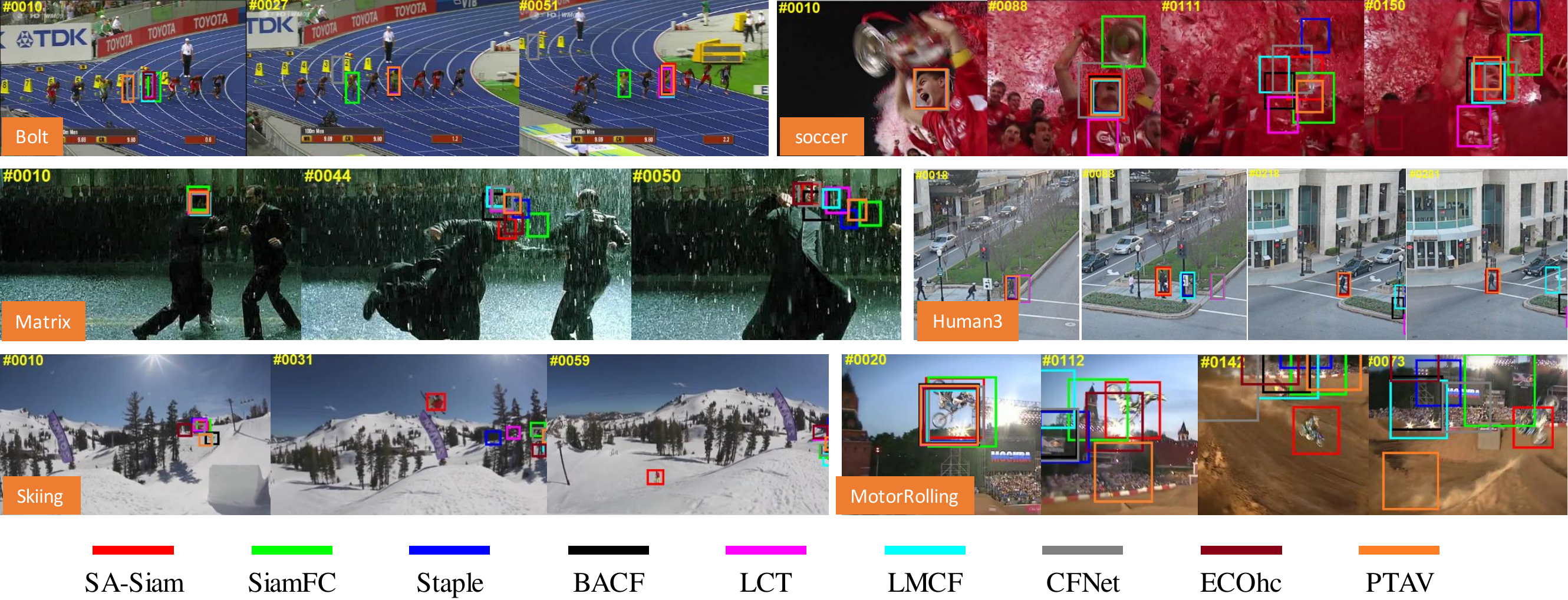} 
    \end{center}
    \caption{Qualitative results comparing SA-Siam with other trackers on sequences \emph{bolt, soccer, matrix, human3, skiing} and \emph{motorrolling}. SA-Siam tracks accurately and robustly in these hard cases. In the very challenging \emph{skiing} and \emph{motorrolling} sequence, SA-Siam can always track to the target when most of other trackers fail.}
    \vspace{-10pt}
    \label{fig:qualitative}
\end{figure*}

\subsection{Comparison with State-of-the-Arts}

We compare SA-Siam with the state-of-the-art real-time trackers on both OTB and VOT benchmarks. Conventionally, a tracking speed beyond 25fps is considered real-time. Our tracker runs at 50fps.

\textbf{OTB benchmarks:} 
We compare SA-Siam with BACF\cite{BACF}, PTAV\cite{PTAV}, DSiamM\cite{DSiam}, EAST\cite{EAST}, Staple\cite{STAPLE} ,SiamFC\cite{SiamFC}, CFNet\cite{CFNET}, LMCF\cite{LMCF} and LCT\cite{LCT} on OTB 2013/50/100 benchmarks. The precision plots and success plots of one path evaluation (OPE) are shown in Fig.\ref{fig:OPEplots}. More results are summarized in Table \ref{tab:res-soa}. The comparison shows that SA-Siam achieves the best performance among these real-time trackers on all three OTB benchmarks. 

Note that the plots in Fig.\ref{fig:OPEplots} are obtained by running the OTB evaluation kit on the tracking result file released by authors. The numbers for BACF \cite{BACF} is different from the paper because the authors only evaluate 50 videos (no \emph{jogging-2}) for OTB-2013. Also, they use per-video average instead of per-frame average to compute the numbers.


\begin{table}
    \begin{center}
    \small{
    \begin{tabular}{l|c c c c}
        \hline
         Tracker   & A & R  & EAO & FPS \\
         \hline
         MDNet     & 0.60 & 0.69 & 0.38 & 1\\
         DeepSRDCF & 0.56 & 1.05 & 0.32 & $<1$\\
         EBT       & 0.47 & 1.02 & 0.31 & 4.4\\
         SRDCF     & 0.56 & 1.24 & 0.29 & 5\\
         \hline
         BACF      & 0.59 & 1.56 & -    & 35\\
         EAST      & 0.57 & 1.03 & 0.34 & 159\\
         Staple	   & 0.57 & 1.39 & 0.30 & 80\\
         SiamFC    & 0.55 & 1.58 & 0.29 & 86 \\
         \hline
         \textbf{SA-Siam}& \emph{0.59} & \emph{1.26} & \emph{0.31} & \emph{50} \\
         \hline
    \end{tabular}}
    \end{center}
    \caption{Comparisons between SA-Siam and state-of-the-art real-time trackers on VOT2015 benchmark. Four top non-real-time trackers in the challenge are also included as a reference.}
    \vspace{-10pt}
    \label{tab:res-vot15}
\end{table}

\textbf{VOT2015 benchmark:} 
VOT2015 is one of the most popular object tracking benchmarks. Therefore, we also report the performance of some of the best non-real-time trackers as a reference, including MDNet\cite{MDNET}, DeepSRDCF\cite{DeepSRDCF}, SRDCF\cite{SRDCF} and EBT\cite{EBT}. Also, SA-Siam is compared with BACF\cite{BACF}, EAST\cite{EAST}, Staple\cite{STAPLE} and SiamFC\cite{SiamFC}.
Table~\ref{tab:res-vot15} shows the raw scores generated from VOT toolkit and the speed (FPS) of each tracker. SA-Siam achieves the highest accuracy among all real-time trackers. While BACF has the same accuracy as SA-Siam, SA-Siam is more robust.

\textbf{VOT2016 benchmark:} 
We compare SA-Siam with the top real-time trackers in VOT2016 challenge \cite{VOT2016} and Staple\cite{STAPLE} , SiamFC\cite{SiamFC} and ECOhc\cite{ECO}. 
The results are shown in Table~\ref{tab:res-vot16}. On this benchmark, SA-Siam appears to be the most robust real-time tracker. The A and EAO scores are also among the top three. 

\begin{table}
    \begin{center}
    \small{
    \begin{tabular}{l|c c c c}
        \hline
         Tracker   & A & R  & EAO & FPS \\
         \hline
         ECOhc & 0.54 & 1.19 & 0.3221 & 60\\
         Staple & 0.54 & 1.42 & 0.2952 & 80\\
         STAPLE+ & 0.55 & 1.31 & 0.2862& $>25$\\
         SSKCF & 0.54 & 1.43 & 0.2771& $>25$\\
         SiamRN & 0.55 & 1.36 & 0.2766& $>25$\\
         DPT & 0.49 & 1.85 & 0.2358& $>25$\\
         SiamAN & 0.53 & 1.91 & 0.2352 & 86\\
         NSAMF & 0.5 & 1.25 & 0.2269& $>25$\\
         CCCT & 0.44 & 1.84 & 0.2230& $>25$\\
         GCF & 0.51 & 1.57 & 0.2179& $>25$\\
         
         \hline
         \textbf{SA-Siam}& \emph{0.54} & \emph{1.08} & \emph{0.2911} & \emph{50}\\
         \hline
    \end{tabular}}
    \end{center}
    \caption{Comparisons between SA-Siam and state-of-the-art real-time trackers on VOT2016 benchmark. Raw scores generated from VOT toolkit are listed. }
    \vspace{-10pt}
    \label{tab:res-vot16}
\end{table}

\textbf{VOT2017 benchmark:}
Table \ref{tab:res-vot17} shows the comparison of SA-Siam with ECOhc\cite{ECO}, CSRDCF++\cite{CSRDCF}, UCT\cite{UCT}, SiamFC\cite{SiamFC} and Staple\cite{STAPLE} on the VOT2017 benchmark. Different trackers have different advantages, but SA-Siam is always among the top tier over all the evaluation metrics. 

\begin{table}
    \begin{center}
    \small{
    \begin{tabular}{l|c c c c}
        \hline
         Tracker    & A & R  & EAO & FPS \\
         \hline  

         SiamDCF & 0.500& 0.473& 0.249  & 60\\
         ECOhc & 0.494& 0.435& 0.238  & 60\\
         CSRDCF++ & 0.453& 0.370& 0.229 &$>25$\\
         CSRDCFf & 0.479& 0.384& 0.227 &$>25$\\
         UCT & 0.490& 0.482& 0.206  & 41\\
         ATLAS & 0.488& 0.595& 0.195 &$>25$\\
         SiamFC & 0.502& 0.585& 0.188  & 86\\
         SAPKLTF & 0.482& 0.581& 0.184 & $>25$\\
         Staple & 0.530& 0.688& 0.169  & 80\\
         ASMS & 0.494& 0.623& 0.169 &$>25$\\
         \hline
         \textbf{SA-Siam} & \emph{0.500} & \emph{0.459} & \emph{0.236} & \emph{50}\\
         \hline
    \end{tabular}} 
    \end{center}
    \caption{Comparisons between SA-Siam and state-of-the-art real-time trackers on VOT2017 benchmark. Accuracy, normalized weighted mean of robustness score, EAO and speed (FPS) are listed. }
    \vspace{-10pt}
    \label{tab:res-vot17}
\end{table}

More qualitative results are given in Fig. \ref{fig:qualitative}. In the very challenging \emph{motorrolling} and \emph{skiing} sequence, SA-Siam is able to track correctly while most of others fail.

\section{Conclusion} \label{sec:conclusion}

In this paper, we have presented the design of a twofold Siamese network for real-time object tracking. We make use of the complementary semantic features and appearance features, but do not fuse them at early stage. The resulting tracker greatly benefits from the heterogeneity of the two branches. In addition, we have designed a channel attention module to achieve target adaptation. As a result, the proposed SA-Siam outperforms other real-time trackers by a large margin on the OTB benchmarks. It also performs favorably on the series of VOT benchmarks. In the future, we plan to continue exploring the effective fusion of deep features in object tracking task.  

\section*{Acknowledgement}

This work was supported in part by National Key Research and Development Program of China 2017YFB1002203, NSFC No.61572451, No.61390514, and No.61632019, Youth Innovation Promotion Association CAS CX2100060016, and Fok Ying Tung Education Foundation WF2100060004.

The authors would like to thank Wenxuan Xie and Bi Li for helpful discussions. 

{\small
\bibliographystyle{ieee}
\bibliography{sasiam_bib}

\begin{thebibliography}{10}\itemsep=-1pt

\bibitem{Tensorflow}
M.~Abadi, A.~Agarwal, P.~Barham, E.~Brevdo, Z.~Chen, C.~Citro, G.~S. Corrado,
  A.~Davis, J.~Dean, M.~Devin, et~al.
\newblock Tensorflow: Large-scale machine learning on heterogeneous distributed
  systems.
\newblock {\em arXiv preprint arXiv:1603.04467}, 2016.

\bibitem{STAPLE}
L.~Bertinetto, J.~Valmadre, S.~Golodetz, O.~Miksik, and P.~H. Torr.
\newblock Staple: Complementary learners for real-time tracking.
\newblock In {\em Proceedings of the IEEE Conference on Computer Vision and
  Pattern Recognition}, pages 1401--1409, 2016.

\bibitem{SiamFC}
L.~Bertinetto, J.~Valmadre, J.~F. Henriques, A.~Vedaldi, and P.~H. Torr.
\newblock Fully-convolutional siamese networks for object tracking.
\newblock In {\em European Conference on Computer Vision Workshop}, pages
  850--865. Springer, 2016.

\bibitem{SCT}
J.~Choi, H.~Jin~Chang, J.~Jeong, Y.~Demiris, and J.~Young~Choi.
\newblock Visual tracking using attention-modulated disintegration and
  integration.
\newblock In {\em Proceedings of the IEEE Conference on Computer Vision and
  Pattern Recognition}, pages 4321--4330, 2016.

\bibitem{ACFN}
J.~Choi, H.~Jin~Chang, S.~Yun, T.~Fischer, Y.~Demiris, and J.~Young~Choi.
\newblock Attentional correlation filter network for adaptive visual tracking.
\newblock In {\em The IEEE Conference on Computer Vision and Pattern
  Recognition (CVPR)}, July 2017.

\bibitem{ECO}
M.~Danelljan, G.~Bhat, F.~Shahbaz~Khan, and M.~Felsberg.
\newblock Eco: Efficient convolution operators for tracking.
\newblock In {\em The IEEE Conference on Computer Vision and Pattern
  Recognition (CVPR)}, July 2017.

\bibitem{DeepSRDCF}
M.~Danelljan, G.~Hager, F.~Shahbaz~Khan, and M.~Felsberg.
\newblock Convolutional features for correlation filter based visual tracking.
\newblock In {\em Proceedings of the IEEE International Conference on Computer
  Vision Workshops}, pages 58--66, 2015.

\bibitem{SRDCF}
M.~Danelljan, G.~Hager, F.~Shahbaz~Khan, and M.~Felsberg.
\newblock Learning spatially regularized correlation filters for visual
  tracking.
\newblock In {\em Proceedings of the IEEE International Conference on Computer
  Vision}, pages 4310--4318, 2015.

\bibitem{CCOT}
M.~Danelljan, A.~Robinson, F.~S. Khan, and M.~Felsberg.
\newblock Beyond correlation filters: Learning continuous convolution operators
  for visual tracking.
\newblock In {\em European Conference on Computer Vision}, pages 472--488.
  Springer, 2016.

\bibitem{PTAV}
H.~Fan and H.~Ling.
\newblock Parallel tracking and verifying: A framework for real-time and high
  accuracy visual tracking.
\newblock In {\em The IEEE International Conference on Computer Vision (ICCV)},
  Oct 2017.

\bibitem{DSiam}
Q.~Guo, W.~Feng, C.~Zhou, R.~Huang, L.~Wan, and S.~Wang.
\newblock Learning dynamic siamese network for visual object tracking.
\newblock In {\em The IEEE International Conference on Computer Vision (ICCV)},
  Oct 2017.

\bibitem{BranchOut}
B.~Han, J.~Sim, and H.~Adam.
\newblock Branchout: Regularization for online ensemble tracking with
  convolutional neural networks.
\newblock In {\em The IEEE Conference on Computer Vision and Pattern
  Recognition (CVPR)}, July 2017.

\bibitem{GOTURN}
D.~Held, S.~Thrun, and S.~Savarese.
\newblock Learning to track at 100 fps with deep regression networks.
\newblock In {\em European Conference on Computer Vision}, pages 749--765.
  Springer, 2016.

\bibitem{SENET}
J.~Hu, L.~Shen, and G.~Sun.
\newblock Squeeze-and-excitation networks.
\newblock {\em arXiv preprint arXiv:1709.01507}, 2017.

\bibitem{EAST}
C.~Huang, S.~Lucey, and D.~Ramanan.
\newblock Learning policies for adaptive tracking with deep feature cascades.
\newblock In {\em The IEEE International Conference on Computer Vision (ICCV)},
  Oct 2017.

\bibitem{RATM}
S.~E. Kahou, V.~Michalski, and R.~Memisevic.
\newblock Ratm: recurrent attentive tracking model.
\newblock {\em arXiv preprint arXiv:1510.08660}, 2015.

\bibitem{BACF}
H.~Kiani~Galoogahi, A.~Fagg, and S.~Lucey.
\newblock Learning background-aware correlation filters for visual tracking.
\newblock In {\em The IEEE International Conference on Computer Vision (ICCV)},
  Oct 2017.

\bibitem{HART}
A.~R. Kosiorek, A.~Bewley, and I.~Posner.
\newblock Hierarchical attentive recurrent tracking.
\newblock {\em arXiv preprint arXiv:1706.09262}, 2017.

\bibitem{VOT2015}
M.~Kristan and et~al.
\newblock The visual object tracking vot2015 challenge results.
\newblock In {\em Proceedings of the IEEE international conference on computer
  vision workshops}, 2015.

\bibitem{VOT2016}
M.~Kristan and et~al.
\newblock The visual object tracking vot2016 challenge results.
\newblock In {\em European Conference on Computer Vision Workshop}, 2016.

\bibitem{VOT2017}
M.~Kristan and et~al.
\newblock The visual object tracking vot2015 challenge results.
\newblock In {\em Proceedings of the IEEE international conference on computer
  vision workshops}, 2017.

\bibitem{ALEXNET}
A.~Krizhevsky, I.~Sutskever, and G.~E. Hinton.
\newblock Imagenet classification with deep convolutional neural networks.
\newblock In {\em Advances in neural information processing systems}, pages
  1097--1105, 2012.

\bibitem{SiamRep}
Y.~Li, X.~Tian, X.~Shen, and D.~Tao.
\newblock Classification and representation joint learning via deep networks.
\newblock In {\em Proceedings of the 26th International Joint Conference on
  Artificial Intelligence}, pages 2215--2221. AAAI Press, 2017.

\bibitem{TC128}
P.~Liang, E.~Blasch, and H.~Ling.
\newblock Encoding color information for visual tracking: Algorithms and
  benchmark.
\newblock {\em IEEE Transactions on Image Processing}, 24(12):5630--5644, 2015.

\bibitem{CSRDCF}
A.~Lukezic, T.~Vojir, L.~Cehovin~Zajc, J.~Matas, and M.~Kristan.
\newblock Discriminative correlation filter with channel and spatial
  reliability.
\newblock In {\em The IEEE Conference on Computer Vision and Pattern
  Recognition (CVPR)}, July 2017.

\bibitem{HCF}
C.~Ma, J.-B. Huang, X.~Yang, and M.-H. Yang.
\newblock Hierarchical convolutional features for visual tracking.
\newblock In {\em Proceedings of the IEEE International Conference on Computer
  Vision}, pages 3074--3082, 2015.

\bibitem{LCT}
C.~Ma, X.~Yang, C.~Zhang, and M.-H. Yang.
\newblock Long-term correlation tracking.
\newblock In {\em Proceedings of the IEEE Conference on Computer Vision and
  Pattern Recognition}, pages 5388--5396, 2015.

\bibitem{TCNN}
H.~Nam, M.~Baek, and B.~Han.
\newblock Modeling and propagating cnns in a tree structure for visual
  tracking.
\newblock {\em arXiv preprint arXiv:1608.07242}, 2016.

\bibitem{MDNET}
H.~Nam and B.~Han.
\newblock Learning multi-domain convolutional neural networks for visual
  tracking.
\newblock In {\em Proceedings of the IEEE Conference on Computer Vision and
  Pattern Recognition}, pages 4293--4302, 2016.

\bibitem{HDT}
Y.~Qi, S.~Zhang, L.~Qin, H.~Yao, Q.~Huang, J.~Lim, and M.-H. Yang.
\newblock Hedged deep tracking.
\newblock In {\em Proceedings of the IEEE Conference on Computer Vision and
  Pattern Recognition}, pages 4303--4311, 2016.

\bibitem{ILSVRC}
O.~Russakovsky, J.~Deng, H.~Su, J.~Krause, S.~Satheesh, S.~Ma, Z.~Huang,
  A.~Karpathy, A.~Khosla, M.~Bernstein, et~al.
\newblock Imagenet large scale visual recognition challenge.
\newblock {\em International Journal of Computer Vision}, 115(3):211--252,
  2015.

\bibitem{SINT}
R.~Tao, E.~Gavves, and A.~W. Smeulders.
\newblock Siamese instance search for tracking.
\newblock In {\em Proceedings of the IEEE Conference on Computer Vision and
  Pattern Recognition}, pages 1420--1429, 2016.

\bibitem{STCT}
L.~Wang, W.~Ouyang, X.~Wang, and H.~Lu.
\newblock Stct: Sequentially training convolutional networks for visual
  tracking.
\newblock In {\em Proceedings of the IEEE Conference on Computer Vision and
  Pattern Recognition}, pages 1373--1381, 2016.

\bibitem{LMCF}
M.~Wang, Y.~Liu, and Z.~Huang.
\newblock Large margin object tracking with circulant feature maps.
\newblock In {\em The IEEE Conference on Computer Vision and Pattern
  Recognition (CVPR)}, July 2017.

\bibitem{OTB13}
Y.~Wu, J.~Lim, and M.-H. Yang.
\newblock Online object tracking: A benchmark.
\newblock In {\em Proceedings of the IEEE conference on computer vision and
  pattern recognition}, pages 2411--2418, 2013.

\bibitem{OTB15}
Y.~Wu, J.~Lim, and M.-H. Yang.
\newblock Object tracking benchmark.
\newblock {\em IEEE Transactions on Pattern Analysis and Machine Intelligence},
  37(9):1834--1848, 2015.

\bibitem{CFNET}
H.~Xu, Y.~Gao, F.~Yu, and T.~Darrell.
\newblock End-to-end learning of driving models from large-scale video
  datasets.
\newblock In {\em The IEEE Conference on Computer Vision and Pattern
  Recognition (CVPR)}, July 2017.

\bibitem{RFL}
T.~Yang and A.~B. Chan.
\newblock Recurrent filter learning for visual tracking.
\newblock In {\em The IEEE International Conference on Computer Vision (ICCV)},
  Oct 2017.

\bibitem{EBT}
G.~Zhu, F.~Porikli, and H.~Li.
\newblock Beyond local search: Tracking objects everywhere with
  instance-specific proposals.
\newblock In {\em Proceedings of the IEEE Conference on Computer Vision and
  Pattern Recognition}, pages 943--951, 2016.

\bibitem{UCT}
Z.~Zhu, G.~Huang, W.~Zou, D.~Du, and C.~Huang.
\newblock Uct: Learning unified convolutional networks for real-time visual
  tracking.
\newblock In {\em The IEEE International Conference on Computer Vision (ICCV)},
  Oct 2017.

\end{thebibliography}
}

\end{document}